\documentclass[10pt,twocolumn,letterpaper]{article}

\usepackage[pagenumbers]{cvpr} 
\usepackage{graphicx}

\definecolor{cvprblue}{rgb}{0.21,0.49,0.74}
\usepackage[pagebackref,breaklinks,colorlinks,allcolors=cvprblue]{hyperref}
\usepackage{multirow} 
\usepackage{graphicx}
\def\paperID{12240} 
\def\confName{CVPR}
\def\confYear{2025}

\title{NTR-Gaussian: Nighttime Dynamic Thermal Reconstruction with 4D Gaussian Splatting Based on Thermodynamics}

\author{
Kun Yang\textsuperscript{1,*},
Yuxiang Liu\textsuperscript{2,*},
Zeyu Cui\textsuperscript{2},
Yu Liu\textsuperscript{2,\dag},
Maojun Zhang\textsuperscript{2},
Shen Yan\textsuperscript{2},
Qing Wang\textsuperscript{1,\dag}\\
\vspace{0.5cm} 
{\small
\textsuperscript{1} Northwestern Polytechnical University \quad
\textsuperscript{2} National University of Defense Technology
\vspace{-0.3cm}
}\\
{\small
\url{https://github.com/ykykykykyk/NTR-Gaussian}
}
}

\begin{document}
\twocolumn[{%
\renewcommand\twocolumn[1][]{#1}%
\maketitle
\begin{center}
    \centering
    \includegraphics[width=12cm]{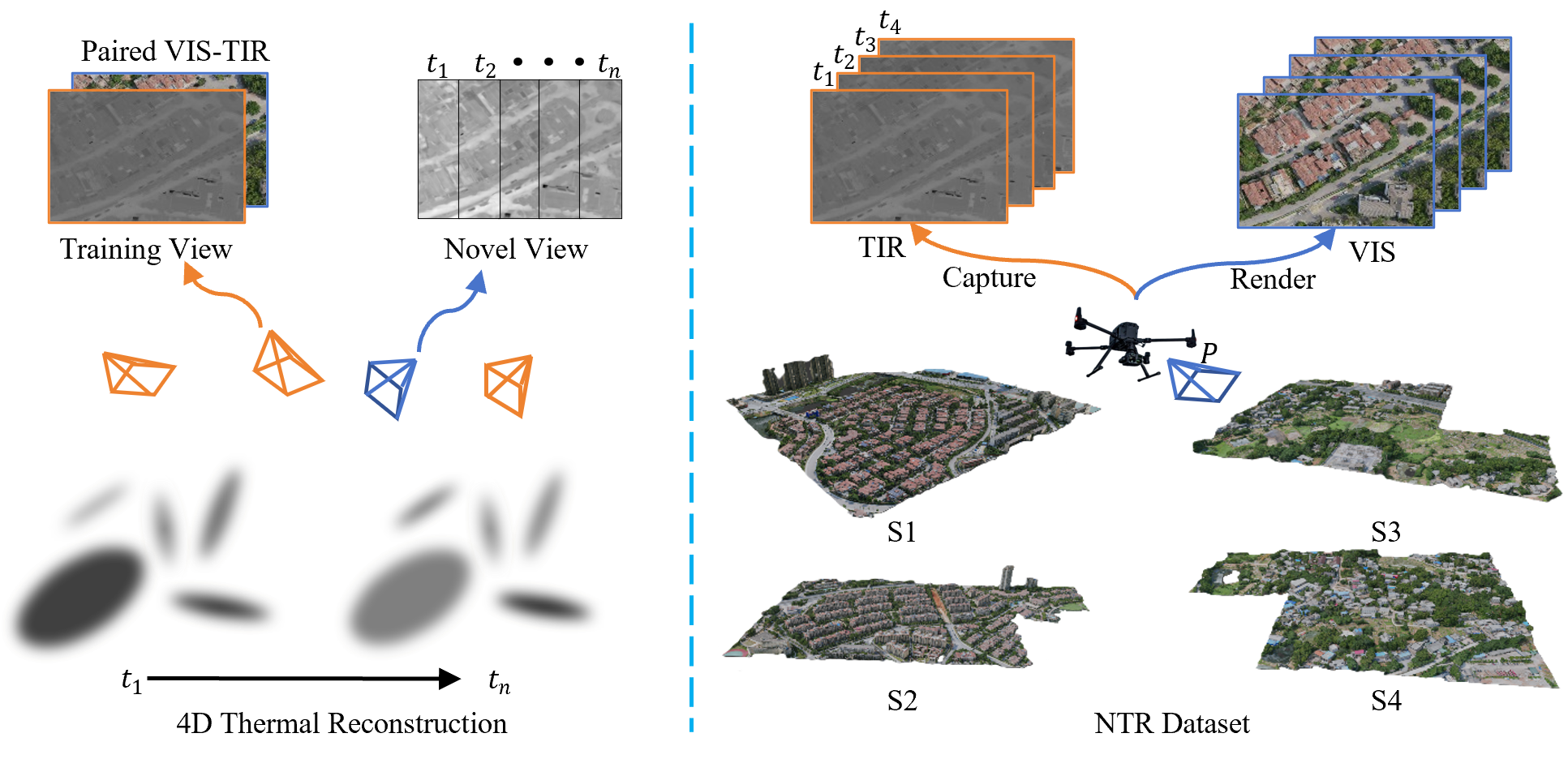} 
    \captionof{figure}{\textbf{Left}: An illustration of dynamic thermal reconstruction based on 3DGS, which learns a dynamic representation of the scene temperature from time-tagged discrete aerial TIR images paired with synthetic RGB images. \textbf{Right}: The proposed NTR dataset contains aerial TIR images with corresponding synthetic RGB images taken at multiple times in four 3D scenes.}
    \label{fig1}
\end{center}%
}]

 
\renewcommand{\thefootnote}{\fnsymbol{footnote}} 
\footnotetext[1]{These authors contributed equally to this work. This research was done when the Kun Yang was in internship at the National University of Defense Technology. \dag Corresponding authors. Email: qwang@nwpu.edu.cn (Qing Wang), jasonyuliu@nudt.edu.cn(Yu Liu)} 

\begin{abstract}
Thermal infrared imaging offers the advantage of all-weather capability, enabling non-intrusive measurement of an object's surface temperature. Consequently, thermal infrared images are employed to reconstruct 3D models that accurately reflect the temperature distribution of a scene, aiding in applications such as building monitoring and energy management. However, existing approaches predominantly focus on static 3D reconstruction for a single time period, overlooking the impact of environmental factors on thermal radiation and failing to predict or analyze temperature variations over time. To address these challenges, we propose the NTR-Gaussian method, which treats temperature as a form of thermal radiation, incorporating elements like convective heat transfer and radiative heat dissipation. Our approach utilizes neural networks to predict thermodynamic parameters such as emissivity, convective heat transfer coefficient, and heat capacity. By integrating these predictions, we can accurately forecast thermal temperatures at various times throughout a nighttime scene. Furthermore, we introduce a dynamic dataset specifically for nighttime thermal imagery. Extensive experiments and evaluations demonstrate that NTR-Gaussian significantly outperforms comparison methods in thermal reconstruction, achieving a predicted temperature error within 1 degree Celsius.
\end{abstract}    
\section{Introduction}
\label{sec:intro}
Thermal imaging technology is a non-contact temperature measurement method that captures heat radiation from objects. The technology has a high tolerance for environmental conditions such as low light and haze, making it widely used in fields like security, building inspection, and energy efficiency analysis~\cite{ThermalReview2022Ramo,vidas2014real,hoegner2018mobile,yang2018fusion}. In recent years, thermal infrared sensors have rapidly advanced, with significant improvements in sensitivity and resolution. Their miniaturization has enabled them to be mounted on drones for large-scale aerial thermal infrared(TIR) image collection, laying the groundwork for the modeling and simulation of large-scale 3D temperature fields. The core objective of 3D thermal scene reconstruction is to reconstruct a three-dimensional scene with thermal properties by integrating thermal images with other visual information. Early approaches to 3D thermal scene reconstruction~\cite{jia20173d,zhao2017real,li2023research,muller2019close,rangel20143d,ham2013automated,javadnejad2017small,vidas2013heatwave,lin2019evaluating,dlesk2021transformations,dlesk2022photogrammetric} primarily followed the strategy of first reconstructing the 3D scene using RGB images and multi-view reconstruction methods~\cite{jadhav2017review}, and then mapping TIR images onto the reconstructed model. However, these methods fail to take full advantage of the unique properties of thermal information and are unable to provide a dynamic representation of temperature change scenarios.

In recent years, several NeRF-based methods have been proposed for reconstructing 3D thermal model using TIR images~\cite{hassan2024thermonerf,ye2024thermal}. However, NeRF is a 3D implicit representation of the scene, which is insufficient in terms of geometric stability. In contrast, 3D Gaussian Splatting~\cite{kerbl20233d} uses a discrete 3D Gaussian distribution to explicitly represent the 3D scene, and the parameters of the 3D Gaussian distribution, such as position, size, and color, are all learnable, making it scalable and easier to integrate with downstream tasks. In 3D thermal reconstruction tasks, some 3DGS-based 3D thermal scene reconstruction methods have also been proposed in recent years~\cite{lu2024thermalgaussian,chen2024thermal3d}. However, these methods usually only perform thermal reconstruction under static conditions for the reconstruction of low-light scenes and rendering of new views, without considering changes in scene temperature, and are unable to dynamically represent and predict temperature variations in the scene.

In order to solve the above problem, we propose NTR-Gaussian based on 3DGS, which takes multi-view TIR images with timestamps and corresponding synthetic visible light (VIS) images as inputs. It directly learns a 4D spatiotemporal representation of the scene's temperature distribution and changes, enabling the rendering of scene temperature maps from new perspectives and at different times (Fig.~\ref{fig1} Left). The scene temperature can be reflected by infrared radiation, integrating convective heat transfer and radiative heat dissipation. Therefore, we utilize a neural network to predict thermodynamic parameters, including emissivity, convective heat transfer coefficient, and heat capacity, enabling NTR-Gaussian to predict the temperature of the scene at each moment during the night through an integration method. In addition, we constructed the NTR dataset specifically for the night-time dynamic thermal reconstruction task for outdoor scenes, containing aerial TIR images for four scenes across multiple time periods(Fig.~\ref{fig1} Right). In summary, our contribution is as follows:
\begin{enumerate}
    \item We constructed the NTR dataset specifically for the night-time dynamic thermal reconstruction.
    \item We propose NTR-Gaussian, which can learn emissivity, convective heat transfer coefficient, and heat capacity to predict the surface temperature of outdoor scenes at night.
    \item We use networks and integration strategies to solve differential equations and improve the accuracy of temperature prediction. Extensive experiments have demonstrated the superiority of our method.
\end{enumerate}

\section{Related Work}

\subsection{3D Thermal Reconstruction}
\label{sec:3dtr}
3D thermal reconstruction is the use of thermal thermal images with 3D reconstruction techniques to reconstruct a 3D scene with thermal properties. High density and high accuracy 3D reconstruction originated from KinectFusion~\cite{newcombe2011kinectfusion} proposed by Newbowed et al. After that there have been also many works in the direction of reconstruction accuracy~\cite{dai2017bundlefusion, schonberger2016structure}, reconstruction efficiency~\cite{niessner2013real, kahler2015very} in this field. In addition to this, some researchers have combined thermal images with 3D reconstruction techniques to reconstruct 3D scenes with temperature information.Rangel et al.~\cite{rangel20143d} used a multimodal system based on a thermal imaging camera and a depth camera to fuse thermal and spatial data to obtain a 3D thermal model. Zhao et al. proposed a real-time handheld 3D temperature field reconstruction system~\cite{zhao2017real}, which improved the accuracy and real-time performance of thermal scene reconstruction by fusing multiple sensor data and calibration techniques. Muller et al. proposed a multimodal fusion T-ICP~\cite{muller2019close} method, which was capable of generating high-fidelity 3D thermal models in real-time based on the robustness of thermal information to changes in the viewpoint and illumination. li et al. proposed an improved stereo matching algorithm~\cite{li2023research}, which effectively improved the quality of 3D reconstruction for thermal thermal stereo vision. However, these traditional methods based on multi-view geometry did not perform as well as the latest deep learning-based methods in rendering from novel views.

NeRF~\cite{mildenhall2021nerf} is considered to be an important milestone in the field of 3D reconstruction due to its ability to render high-fidelity images in new perspectives, and a number of NeRF-based 3D thermal reconstruction methods have been proposed recently.The ThermoNeRF~\cite{hassan2024thermonerf} proposed by Hassan et al. and the Thermal-NeRF~\cite{ye2024thermal} methods proposed by Ye et al. both combine thermal image ensemble with NeRF method. However, limited by the slow rendering speed of NeRF itself and the problem of implicit representation, it is difficult to be practically applied.

\subsection{3D Gaussian Splatting}
\label{sec:3DGS}
\begin{figure*}[!t]
\centering
  \centering
   \includegraphics[width=1.\linewidth]{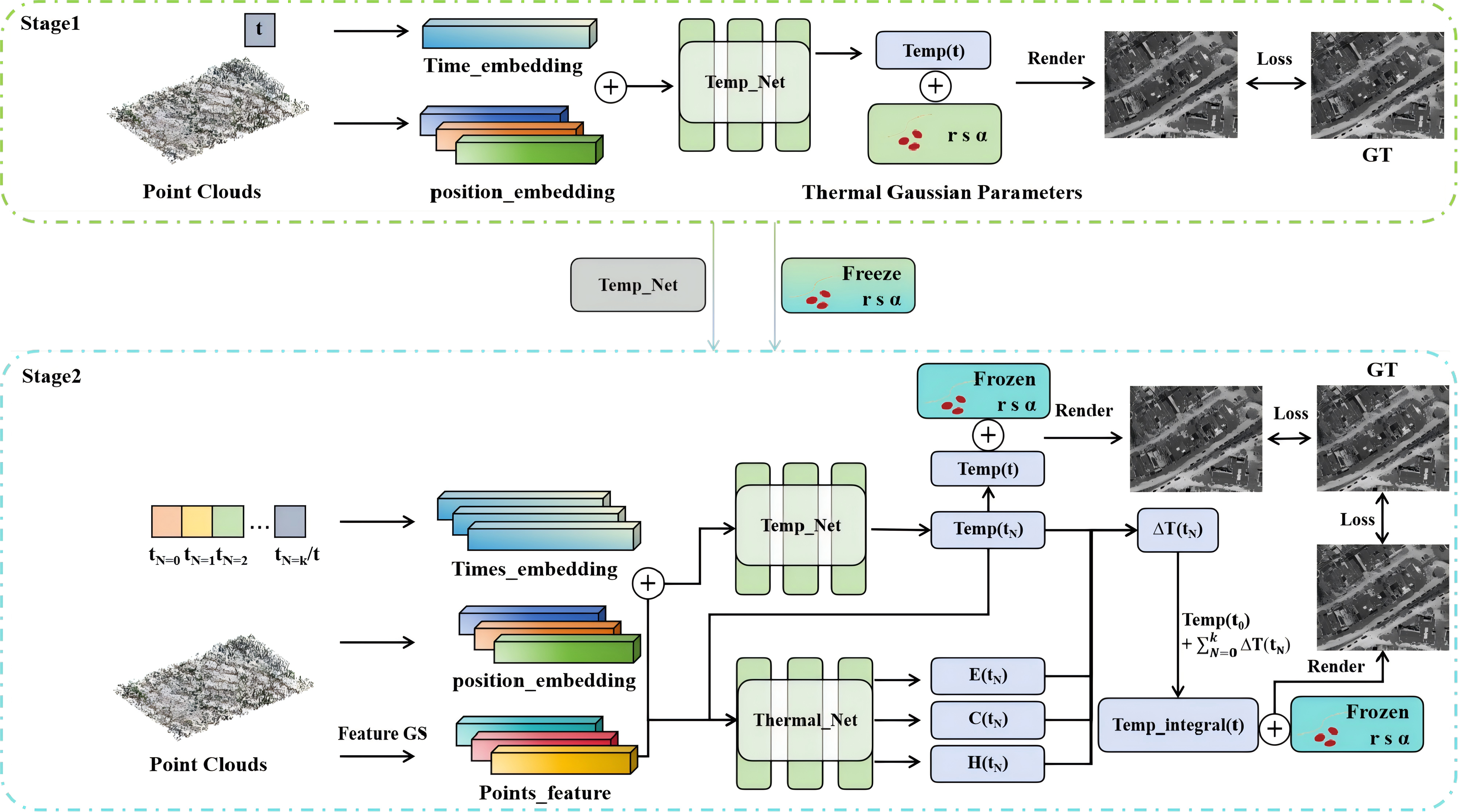}

   \caption{\textbf{Method Overview.} Our method consists of two stages. In the first stage, we input the discrete time encoding from the true thermal temperature image corresponding to time $t$ and the position encoding obtained from the tilted photography model in the collected thermal region. We optimize Temp\_Net, $\alpha$, $r$ and $s$ to obtain the Gaussian sphere infrared radiation temperature Temp($t$) and attribute values corresponding to discrete time. In the second stage, given the initial time $t_0$ and the true thermal temperature image corresponding to time $t$, divide this time interval into $N$ parts to obtain $N$ time encodings. Combine the position encoding through Temp\_Net to obtain Temp($t_N$). Then, use Temp($t_N$), time encodings, position encoding, and semantic feature information obtained from Feature\_GS~\cite{zhou2024feature} through Thermal\_Net to obtain the emissivity E($t_N$), convective heat transfer coefficient C($t_N$), and heat capacity H($t_N$) of the Gaussian sphere for each moment in the corresponding time interval. By applying thermodynamic equations, we calculate the temperature variation $\Delta$T($t_N$) of the Gaussian sphere at each moment and integrate to obtain the temperature Temp\_integral($t$) corresponding to time $t$. We freeze the Gaussian sphere parameters obtained in the first stage and render the thermal prediction images separately from Temp($t$) and Temp\_integral($t$). By jointly optimizing the network, we ensure that these two values are consistent at each moment, allowing us to predict the infrared radiation temperature values at any moment in any nighttime scene based on physical laws.}
   \label{fig2}
\end{figure*}
3D Gaussian Splatting~\cite{kerbl20233d} is a novel and disruptive technique in the field of 3D reconstruction. Unlike methods such as NeRF, 3DGS revolutionizes the representation of 3D scenes by using millions of explicit Gaussian functions. This technique combines neural network-based optimization with the benefits of structured data representation to not only enable photo-realistic rendering from new perspectives, but also significantly enhances real-time rendering and introduces the ability to manipulate and edit 3D scenes. These features enable 3DGS to be highly compatible with a wide range of downstream applications, and are thus considered to be the core foundation of the next generation of 3D reconstruction technology~\cite{dalal2024gaussian}.
Even though 3DGS relies only on RGB modal images during training, its final output generates millions of point cloud-like Gaussian functions. This explicit representation makes 3DGS particularly well suited for multimodal fusion with other devices that directly capture point clouds of field points, such as depth cameras and LiDAR. Studies ~\cite{matsuki2024gaussian,yan2024gs,keetha2024splatam} had successfully integrated a depth camera into the 3DGS approach, enabling simultaneous localization and map construction (SLAM) based on 3DGS. In addition, other studies~\cite{li2024dngaussian,zhu2025fsgs} have combined depth images~\cite{ranftl2020towards} inferred from a pre-trained monocular depth estimation model~\cite{bhat2023zoedepth}  with RGB modalities to dramatically improve the rendering quality and generate more accurate geometric structures.Wu et al. proposed 4DGS~\cite{wu20244d}, which introduces deformation fields on top of 3DGS to model 3D dynamic scenes, and achieves high-quality 3D dynamic scene reconstruction. Besides, there are a few works~\cite{chen2024thermal3d, lu2024thermalgaussian}  that combine thermal images with 3DGS to realize 3DGS-based thermal scene reconstruction, but these methods only perform thermal reconstruction in static space, which restricts the accuracy of the reconstruction results and hinders its wider practical application.

In contrast, our method considers the temperature as a manifestation of infrared radiation, integrates the effects of convective heat transfer and radiative heat dissipation, and utilizes neural networks to predict thermodynamic parameters, including emissivity, convective heat transfer coefficient, and heat capacity. In addition, our method also predicts the thermal temperature of the scene at each moment of the night by integrating, so as to obtain a more accurate and continuous three-dimensional dynamic thermal scene.
\section{NTR Dataset}
To evaluate methods for dynamic thermal 3D reconstruction and novel viewpoint TIR image synthesis, we have developed the NTR dataset as shown in Fig. \ref{fig1}. This dataset covers four distinct scenes: two urban scenes primarily characterized by buildings and roads (S1, S2), and two suburban scenes dominated by farmland and ponds (S3, S4). For each scene, we capture aerial TIR images at four time intervals during the night. 
All TIR images are captured by a DJI Matrice 300 RTK drone\footnote{{\url{https://www.dji.com/cn/matrice-300}}} equipped with a DJI H20T\footnote{{\url{https://www.dji.com/cn/zenmuse-h20-series}}} thermal infrared camera. The flight altitude is set at 250 meters. The thermal imaging sensor has a resolution of $640\times512$, a DFOV of 40.6$^\circ$, a focal length of 13.5 mm, and a temperature range from -40$^\circ$C to 550$^\circ$C. These images effectively capture the temporal variations in thermal radiation from objects, such as buildings, under nighttime conditions, allowing the TIR images with timestamps to serve as representative samples of dynamic temperature changes in the scene.Based on the high-precision 3D texture model provided by the UAV4DL dataset\cite{Wu2023uav4l}, we adopt the render-to-match framework~\cite{yan2023render,liu2024atloc} to provide accurately calibrated poses for all TIR images, and use these poses to render corresponding synthetic RGB images for the TIR images. The detailed generation process of the dataset will be described in the Supplementary Material.
\section{Method}
\label{sec:formatting}

\subsection{Overview}
\label{sec:Overview}
The pipeline of the proposed NTR-Gaussian is shown in Fig.~\ref{fig2}. Our method involves two stages: in the first stage, we optimize Temp\_Net to obtain the Gaussian sphere thermal temperature and attribute values for discrete times based on the discrete time encoding from the ground truth thermal temperature images corresponding to time $t$ and the position encoding obtained from the tilted photography model. 

In the second stage, we divide the time interval from the initial time $t_0$ and the ground truth thermal temperature images corresponding to time $t$ into $N$ parts to obtain $N$ time encodings. Then, based on thermodynamic equations, Temp($t_N$), time encodings, position encoding and semantic features derived from Feature\_GS~\cite{wu20244d} are utilized by Thermal\_Net to calculate the temperature variation $\Delta$T($t_N$) for each time segment within the interval.

By jointly optimizing the network, we could predict the infrared temperature values for any moment in nighttime scenes.

\subsection{4D Gaussian Splatting}
\label{sec:4D}
4D Gaussian Splatting~\cite{wu20244d} is based on 3D Gaussian Splatting~\cite{kerbl20233d}. 3D Gaussian Splatting is an explicit representation of a scene, 4D Gaussian Splatting incorporates a time variable on top of the 3D Gaussian Splatting, enabling certain Gaussian parameters to vary with time.

3D Gaussian Splatting is a technique used for representing and rendering point clouds in three-dimensional space. It creates smooth and continuous 3D representations by applying Gaussian distributions to each point in the point cloud.  The Gaussian distribution, characterized by its bell-shaped curve, is a common probability distribution. In this spatial representation, the Gaussian distribution is expressed as a three-dimensional Gaussian function:
\begin{equation}
\small
G(\mathbf{x}) = \frac{1}{(2\pi)^{3/2} |\boldsymbol{\Sigma}|^{1/2}} \exp \left( -\frac{1}{2} (\mathbf{x} - \boldsymbol{\mu})^\top \boldsymbol{\Sigma}^{-1} (\mathbf{x} - \boldsymbol{\mu}) \right),
\label{eq:1}
\end{equation}
where $\boldsymbol{\mu}$ is the mean vector, representing the center of the Gaussian distribution, and $\boldsymbol{\Sigma}$ is the covariance matrix, describing the shape and orientation of the distribution.

In 3D Gaussian Splatting, each point is represented as a three-dimensional Gaussian distribution. These Gaussians are defined by their mean position and covariance matrix. The covariance matrix is decomposed into $\mathbf{r}$ and $\mathbf{s}$, 
\begin{equation}
\boldsymbol{\Sigma} = \mathbf{r}\mathbf{s}\mathbf{s}^{\top}\mathbf{r}^{\top},
\label{eq:3}
\end{equation}
where $\mathbf{r}$ represents the rotation of the Gaussian ellipsoid, and $\mathbf{s}$ represents its shape.

In 4D Gaussian Splatting, with the introduction of time $t$, it learn the variations in Gaussian sphere positions and covariances at each moment $t$ relative to the initial moment $t_0$, while maintaining color $c$ and opacity constants $\alpha$. This can be expressed as:
\begin{equation}
\Delta \boldsymbol{x},\Delta \boldsymbol{y},\Delta \boldsymbol{z},\Delta \boldsymbol{r},\Delta \boldsymbol{s}= F_{4D}(\boldsymbol{t}),
\end{equation}
Where $\Delta$$\mathbf{x}$, $\Delta$$\mathbf{y}$ and $\Delta$$\mathbf{z}$ represent the variation in Gaussian sphere positions at each moment, while $\Delta$$\mathbf{r}$ and $\Delta$$\mathbf{s}$ denote the variation in covariance.

In the construction of thermal scenes, we replace the color $\mathbf{c}$ with the infrared radiation temperature $\mathbf{T}$.
\subsection{Optimization of the coarse network and Gaussian sphere parameters}
\label{sec:coarse}
\begin{figure}[t]
\centering
  \centering
   \includegraphics[width=1.\columnwidth]{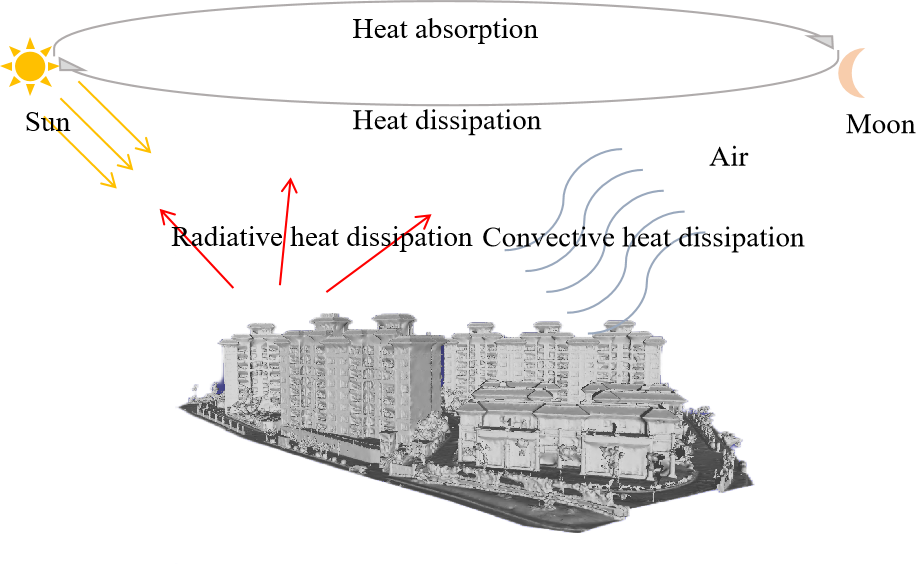}

   \caption{\textbf{Thermodynamic Theory.} The temperature of outdoor objects shows a heat absorption and release performance during the day and night. During the day, the objects absorb heat from solar radiation, while at night, with the setting of the sun, the objects mainly release heat through their own radiation and convective heat exchange.}
   \label{fig3}
\end{figure}
Similar to 4D Gaussian Splatting, in our case, the radiation temperature values change over time in the scene, while the other Gaussian sphere parameters remain constant over time.

\noindent\textbf{Training Process.} In the first stage, we first need to learn a mapping between discrete time and Gaussian sphere temperatures, as well as Gaussian sphere parameters, to ensure that subsequent optimizations can be based on this foundation.

In this process, we utilize a MLP as the Temp\_Net for predicting the infrared radiation temperature of Gaussian spheres Temp($t$). The inputs to the network are the time encoding and Gaussian sphere position encoding. The time encoding corresponds to the time $t$ of the thermal supervised image, while the Gaussian sphere positions are obtained from the point cloud of the corresponding region in the oblique photography model.

The predicted Gaussian sphere temperatures and the Gaussian sphere parameters to be optimized are combined to render the ground truth image on the corresponding camera pose. The loss is then computed. Subsequently, the neural network Temp\_Net and the parameters of the Gaussian sphere are trained and optimized throughout this process.

\noindent\textbf{Loss Function.} In the above optimization process, we utilize L1 loss $L_{Temp}$ to measure the pixel-wise difference between the ground truth image and its corresponding rendering. The $L_{SSIM}$ loss is leveraged to measure the quality loss of the rendered images. The full objective function as well as its components are given as follows:
\begin{equation}
L = \lambda_1L_{Temp} + \lambda_2L_{SSIM},
\label{eq:5}
\end{equation}
where we empirically set $\lambda_1$ = 0.8 and $\lambda_2$ = 0.2.

\subsection{Optimization of thermodynamics-based fine network}
\label{sec:fine}

\begin{table*}[!t]
    \footnotesize
    \centering
    \caption{Quantitative comparison on the scene S1, S2, S3 and S4}
    \resizebox{\textwidth}{!}{%
    \begin{tabular}{lccccccccccccc}
        \toprule
        \multirow{2}{*}{Method} & \multicolumn{3}{c}{S1} & \multicolumn{3}{c}{S2} & \multicolumn{3}{c}{S3} & \multicolumn{3}{c}{S4} \\
        \cmidrule(lr){2-4} \cmidrule(lr){5-7} \cmidrule(lr){8-10} \cmidrule(lr){11-13}
        & PSNR\(\uparrow\)  & SSIM\(\uparrow\)  & MAE\(\downarrow\)  & PSNR \(\uparrow\) & SSIM \(\uparrow\) & MAE \(\downarrow\) & PSNR \(\uparrow\) & SSIM \(\uparrow\) & MAE \(\downarrow\) & PSNR \(\uparrow\) & SSIM \(\uparrow\) & MAE \(\downarrow\) \\
        \midrule    
        3DGS ~\cite{kerbl20233d} & 33.21 & 0.953 & 1.144 & 31.29 & 0.955 & 1.766 & 29.28 & 0.952 & 2.267 & 31.95 & 0.927 & 1.644 \\
        4DGS ~\cite{wu20244d} & 32.19 & 0.961 & 1.655 & 29.17 & 0.961 & 2.475 & 29.78 & 0.954 & 2.224 & 32.06 & 0.958 & 1.616 \\
        Thermal3D-GS ~\cite{chen2024thermal3d} & 34.03 & 0.957 & 1.106 & 30.61 & 0.960 & 1.840 & 33.26 & 0.968 & 1.276 & 31.79 & 0.956 & 1.432 \\
        Ours & \textbf{34.50} & \textbf{0.962} & \textbf{0.947}& \textbf{31.43} & \textbf{0.964} & \textbf{1.709} & \textbf{35.15} & \textbf{0.972} & \textbf{0.959} & \textbf{34.61} & \textbf{0.964} & \textbf{1.065} \\
         \bottomrule
    \end{tabular}
    }
    \label{tab:1}
\end{table*}

\textbf{Principle of Nighttime Heat Dissipation.} The simplified theory of infrared radiation absorption and release based on thermodynamic principles, as shown in Fig.~\ref{fig3}. During the daytime, the thermal temperature of the scene is determined by both absorption and release processes, with absorption mainly from solar radiation and release mainly from convective heat exchange and radiation. At night, the outdoor scene's thermal temperature is mainly determined by the release process, and we focus on the nighttime period. The convective heat exchange and radiative heat release processes are represented by Equ.~\ref{eq:6} and Equ.~\ref{eq:7}, respectively.
\begin{figure}[t]
\centering
\includegraphics[width=1.\columnwidth]{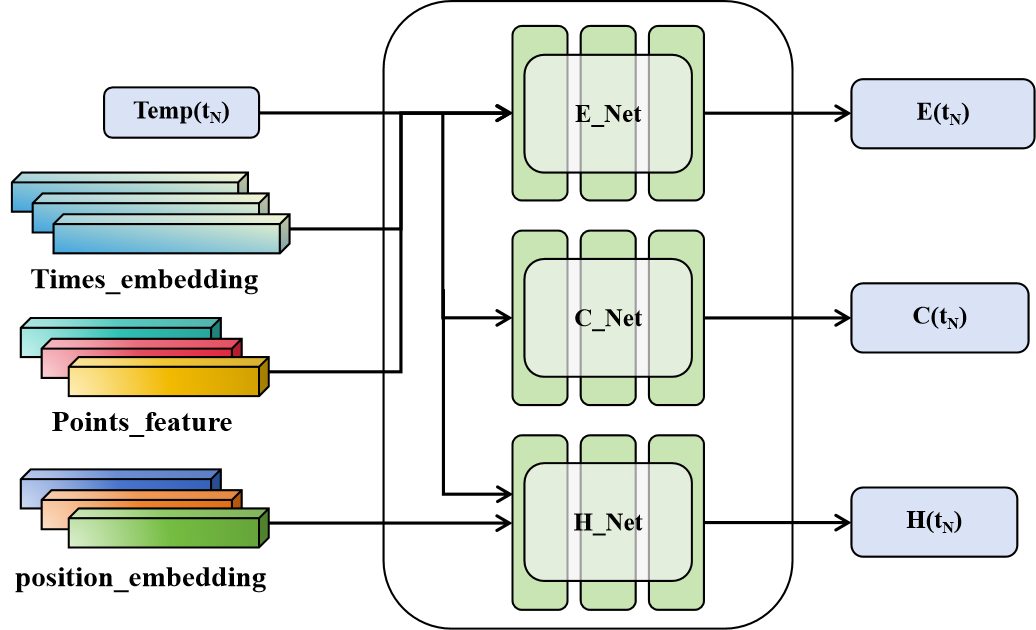} 
\caption{\textbf{The structure of $\mathbf{Thermal\_Net}$.} Consists of three parts, namely E\_Net, C\_Net, and H\_Net. They respectively predict the emissivity, convective heat transfer coefficient, and heat capacity of the target, where the heat capacity is related to the object's position, hence requiring input position encoding.}
\label{fig4}
\end{figure}
\begin{equation}
Q_{con} = C*(T(t) - T_{m}),
\label{eq:6}
\end{equation}
Where $C$ represents the convective heat transfer coefficient of different objects in the scene, which is dependent on the radiative temperature and object's characteristics. $T(t)$ represents the radiative temperature at each moment, and $T_{m}$ represents the environmental temperature at each moment. This formula calculates the amount of heat exchanged between the scene and the air through convection within a unit of time.
\begin{equation}
Q_{rad} = E*\sigma*(T(t))^4,
\label{eq:7}
\end{equation}
Here, $E$ represents the emissivity of different objects in the scene, which is dependent on the radiative temperature and object's characteristics. $\sigma$ is the Stefan-Boltzmann constant, and $T(t)$ represents the radiative temperature at each moment. This formula calculates the amount of heat radiated by the scene within a unit of time.

Therefore, by using the differential equation relationship as shown in Equ.~\ref{eq:8}, solving the differential equation will allow us to calculate the radiative temperature of the object at each moment.
\begin{equation}
T’(t) = (Q_{rad} + Q_{con})/H,
\label{eq:8}
\end{equation}
Where $H$ represents the thermal capacity of different objects, which depends on the object's position and characteristics.
\begin{figure*}[t]
\centering
  \centering
   \includegraphics[width=1.\linewidth]{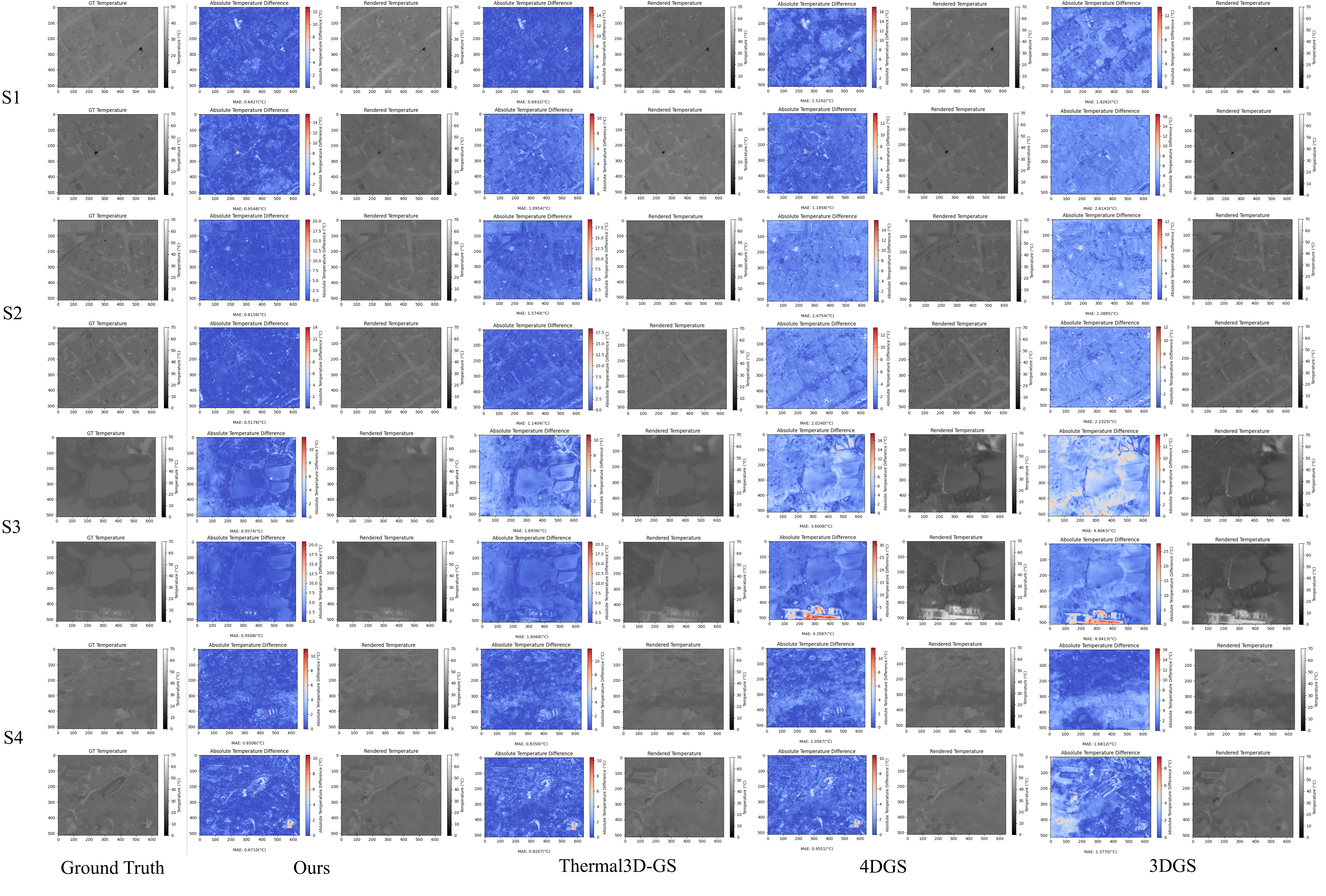}

   \caption{\textbf{Qualitative comparison.} The leftmost column displays the ground truth (GT) temperature images. On the right side of the dividing line, for each method, the right column shows the predicted temperature images, while the left column visualizes the error between the ground truth and predicted temperatures. The error has been normalized using the min-max normalization method and mapped to a red-blue color scale, where regions closer to red represent larger absolute temperature errors, and regions closer to blue represent smaller errors. Our method outperforms other methods in the distribution of temperature differences across the entire scene and in terms of Mean Absolute Error (MAE).}
   \label{fig5}
\end{figure*}
\begin{figure*}[t]
  \centering
  \includegraphics[width=1\linewidth]{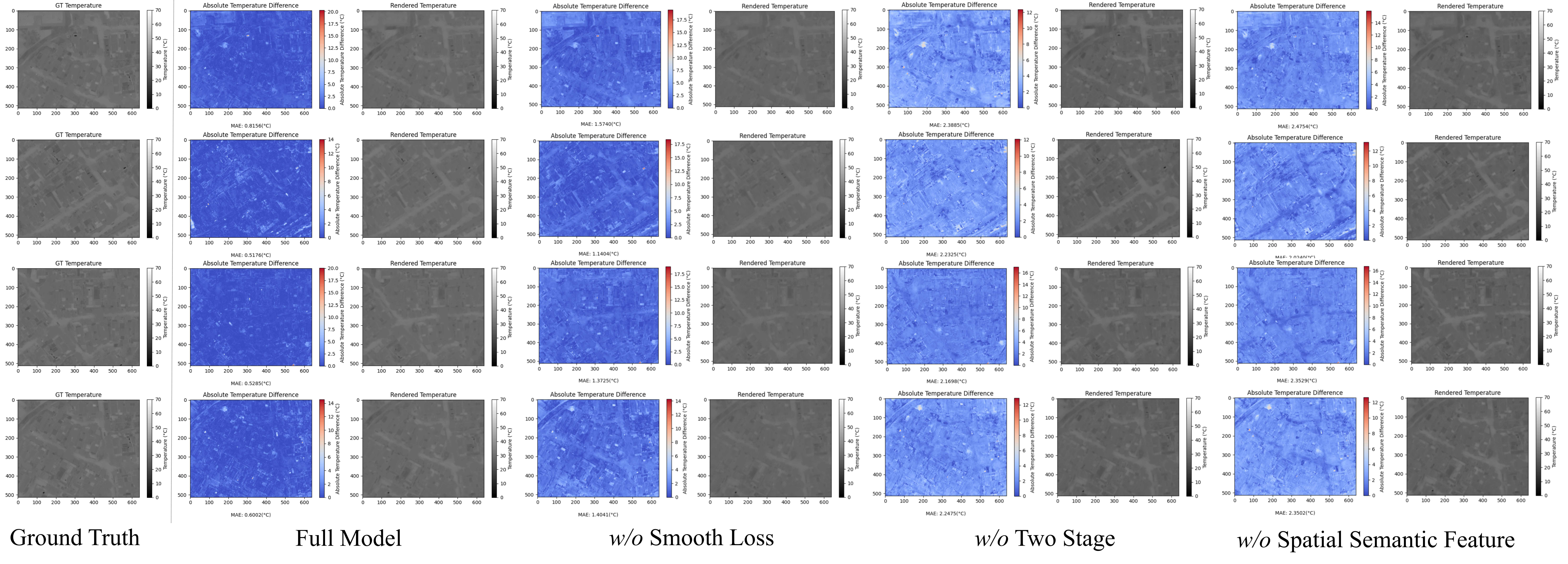}
  \caption{\textbf{Qualitative ablation study on smooth loss, two stages and spatial semantic feature.} We use image Mean Absolute Error (MAE) to measure the accuracy of temperature prediction. Visual comparisons are conducted separately in the complete model, without using two-stage, smooth loss, and spatial semantic features. Experimental results show that our strategy is optimal.}
  \label{fig6}
\end{figure*}

\noindent\textbf{Solving Fine Radiative Temperature.}  To solve the numerical solution of the differential equation, we use the Temp\_Net trained in stage one as $T(t)$ in the differential equation. Then, we introduce an MLP as the Thermal\_Net for predicting thermophysical quantities. As shown in the Fig.~\ref{fig4}, it is the structure of the Thermal\_Net. The factors influencing the thermophysical quantities include the temperature  $T(t)$ of each Gaussian sphere obtained from Temp\_Net, the semantic features of each Gaussian sphere obtained from Feature\_GS, time encodings divided into N segments from time $T_{0}$ to $t/t_{N=k}$ and position encoding of the Gaussian spheres obtained from the tilt model. The output provides the corresponding physical quantities $E(t)$, $H(t)$, $C(t)$ for each Gaussian sphere. Then, using Equ.~\ref{eq:8}, we calculate the rate of temperature change at each moment, i.e., T'(t) in the differential equation.

By freezing the Gaussian sphere parameters obtained in the first stage and combining $T_{integral}(t_{N=k})$ obtained from Equ.~\ref{eq:9} with $T(t_{N=k})$ obtained from Temp\_Net, we render the temperature imgae. We jointly optimize the network with the ground truth temperature map to ensure the consistency of $T_{integral}(t_{N=k})$ and $T(t_{N=k})$. After training is completed, it ensures that we can integrate to obtain the infrared radiation temperature values of any scene at any time.
\begin{equation}
T_{integral}(t_{N=k}) = T(t_{0}) + \sum_{n=1}^{k}\Delta T(t_{N}).
\label{eq:9}
\end{equation}

\noindent\textbf{Loss Function.} In addition to the loss in stage one, we also added a smoothing loss term to ensure that there are no abrupt changes in temperature within the time interval and to guarantee smooth variations. The loss function is shown in Equ.~\ref{eq:10}.
\begin{equation}
L = \lambda_1L_{Temp} + \lambda_2L_{SSIM} + \lambda_3L_{Smooth},
\label{eq:10}
\end{equation}
where we empirically set $\lambda_1$ = 0.8 , $\lambda_2$ = 0.2 and $\lambda_2$ = 10.
\section{Experiments}
\subsection{Experiment Setup}
\label{sec:Setup}
\noindent\textbf{Data division.} The TIR images in our NTR dataset for each scene are captured across four time periods. We use the first and last time periods for supervised training, while the intermediate time periods serve as the test set. Additionally, we normalize the infrared thermal data into temperature maps. There are a total of 3052 TIR images, which we divided into training and testing sets in a one-to-one ratio.

\noindent\textbf{Evaluation Metrics.}  Following the evaluation metrics in 3DGS~\cite{kerbl20233d} and temperature evaluation metrics, we use PSNR, SSIM~\cite{wang2004image} and Mean Absolute Error (MAE) of temperature values as the evaluation metrics.

\noindent\textbf{Implementation Details.} During the training process, we downsample the original point clouds of each scene to ensure spatial efficiency, iterating 20K times in both stage one and stage two. Our experimental environment consists of a dual NVIDIA RTX 4090 GPU (24 GB memory), with image resolution set at 640$\times$512 for both training and testing. More details, including the dataset used, training settings, and additional information, can be found in our supplementary materials.
\subsection{Comparison experiment}
\label{sec:Comparisons} 

\noindent\textbf{Comparison Methods}. We compared our method dynamically with the 4D Gaussian Splatting~\cite{wu20244d} method, and compared it with Thermal Gaussian Splatting~\cite{chen2024thermal3d} and 3D Gaussian Splatting~\cite{kerbl20233d} to evaluate the reconstruction effect of our static moments. For the sake of fairness, we conduct training and testing on the same dataset with the same settings. When comparing with static methods, we don't use time as an input to prove that static methods can only learn the average temperature and cannot obtain accurate temperature values. When comparing with dynamic methods, we use the same temperature for input to prove that a simple temperature mapping is not accurate enough, and our method is more in line with the physical principles.

\noindent\textbf{Results}. As shown in Tab.~\ref{tab:1}, our NTR-Gaussian outperforms other approaches in quantitative metrics, achieving a predicted temperature error within 1 degree Celsius. The Fig.~\ref{fig5} presents a visual comparison of the Mean Absolute Error (MAE) in temperature. It can be observed that our method produces more accurate temperature predictions. It is important to note that the relatively small difference between the metrics of static and dynamic methods stems from the fact that static methods learn the average behavior over a certain period. As a result, static methods may perform better at specific moments but worse at others.
\subsection{Ablation Study}
\label{sec:Ablation}
\begin{figure}[t]
\centering
\includegraphics[width=1.\columnwidth]{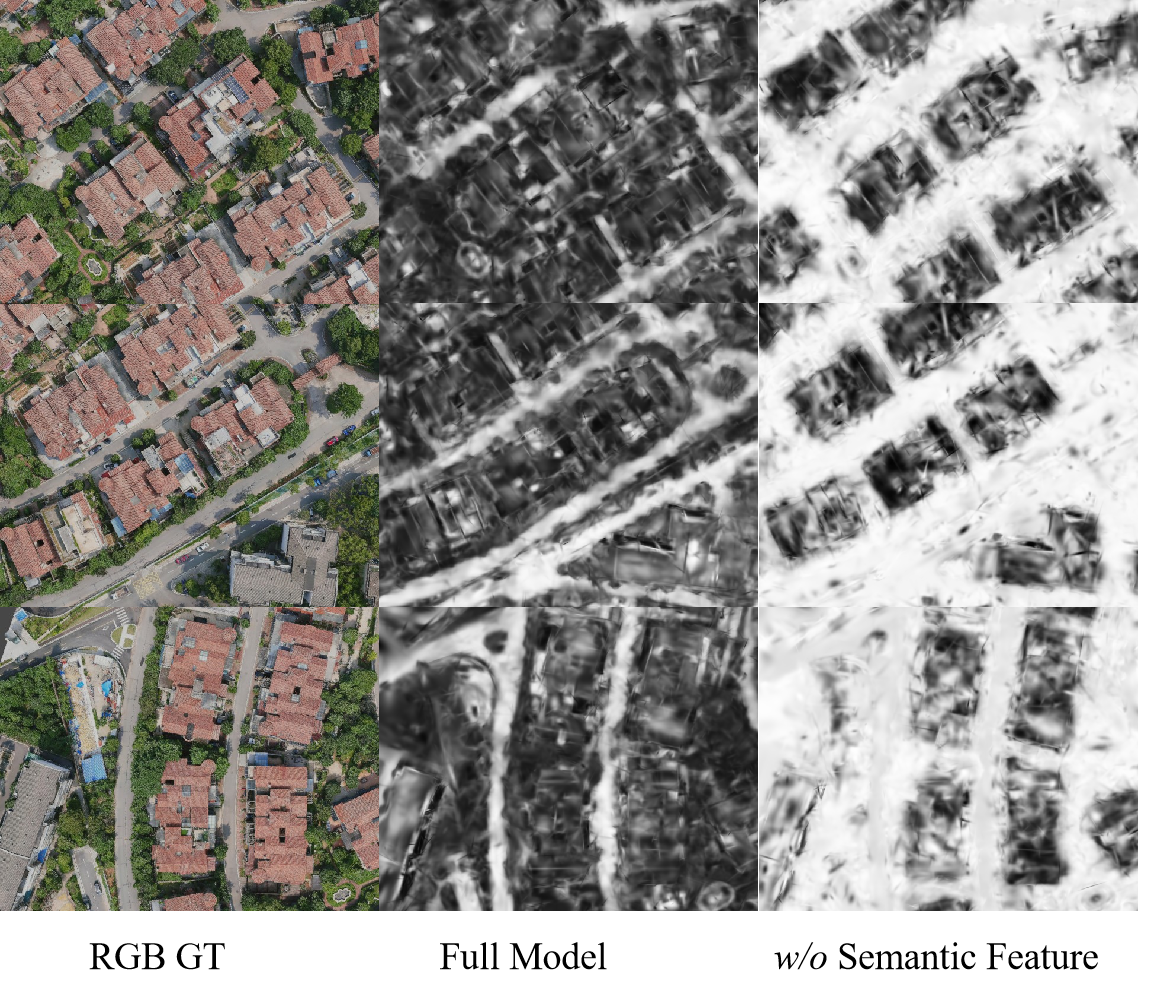} 
\caption{The presence of spatial semantic features has an impact on the prediction of physical quantities. Taking emissivity as an example, when spatial semantic features are absent, many details of the physical quantities are poor, which in turn affects the prediction of temperature.}
\label{fig7}
\end{figure}
We conduct ablation experiments from three perspectives to demonstrate the effectiveness of each component of our method: whether to use two-stage training, whether to use spatial semantic features, and whether to use the Smoothing loss term. Using S1 as an example, we present the quantitative results of these three ablation experiments in Tab.~\ref{tab:2}.

\noindent\textbf{Two Stage Pipeline.} We test a single-stage training strategy, in which these networks and parameters were trained in one phase. We found that in the test results, PSNR, SSIM, and temperature MAE were all lower than when using the two-stage training approach. Table~\ref{tab:2} presents the quantitative metrics for both strategies, and in Fig.~\ref{fig6}, the two-stage strategy demonstrates more accurate temperature predictions for unknown time points.

\noindent\textbf{Smoothing Loss Term.} We conducted experiments without using the smoothness loss. Since, in principle, temperature changes over time are smooth and continuous, as shown in Tab.~\ref{tab:2}, temperature predictions at specific locations and times were naturally less accurate than when the smoothness loss was applied. And as shown in Fig.~\ref{fig6}, the average temperature prediction is more accurate when the smoothing loss is used.
\begin{table}[t]
    \centering
    \caption{\textbf{Quantitative ablation study.} We report PSNR, SSIM, and MAE for evaluating the rendering quality.}
    \vspace{-2mm}
    \resizebox{\columnwidth}{!}{%
    \begin{tabular}{lccc}
        \toprule
        Model & PSNR$\uparrow$  & SSIM$\uparrow$  & MAE$\downarrow$  \\
        \midrule
        {w/o} Smoothing Loss & 34.21 & 0.945 & 1.270 \\
        {w/o} Spatial Semantic Feature & 32.80 & 0.934 & 1.498  \\
        {w/o} Two Stages & 32.93 & 0.932 & 1.496  \\
        Full Model & \textbf{34.50} & \textbf{0.962} & \textbf{0.947} \\
        \bottomrule
    \end{tabular}%
    }
    \label{tab:2}
\end{table}

\noindent\textbf{Spatial Semantic Features.} We conducted experiments without using spatial semantic features. As shown in Tab.~\ref{tab:2}, the quantitative metrics in our test results were all lower compared to when spatial semantic features were used. Additionally, in the prediction process of physical quantities, these values appeared disorganized and unrealistic without spatial semantic features. As shown in Fig.~\ref{fig7}, using spatial semantic features significantly improves the accuracy of physical quantity predictions compared to not using them.

\section{Conclusion}
In conclusion, we present the NTR-Gaussian method as a novel approach to overcoming the limitations of existing static thermal 3D reconstruction by incorporating dynamic environmental factors and temporal information. Our method leverages the capabilities of neural networks to predict essential thermodynamic parameters, such as emissivity, convective heat transfer coefficient, and heat capacity, thereby enabling the accurate forecasting of temperature variations across nighttime scenes. Furthermore, the proposed NTR dataset comprising aerial TIR images of multiple outdoor scenes across different time periods, serves as a valuable benchmark for future research.

The incorporation of multi-view TIR images with timestamps, combined with synthetic VIS images, enables the creation of a comprehensive 4D spatiotemporal representation of temperature distribution and variation. This advancement not only improves the accuracy of thermal reconstruction but also broadens its applicability to dynamic environmental conditions, which are critical for applications such as building monitoring and energy management. Our experimental results consistently demonstrate that NTR-Gaussian outperforms comparative methods, achieving superior accuracy in thermal reconstruction. However, our method still has limitations, primarily in two areas. First, it requires individualized optimization for each scene and exhibits limited generalization performance, making it challenging to apply uniformly across different scenes and seasons. Second, the temperature prediction does not incorporate additional environmental constraints, such as humidity and wind speed. Addressing these issues will be crucial for the broader application of dynamic thermal temperature prediction in the future.

{
    \small
    \bibliographystyle{ieeenat_fullname}
    \bibliography{main}
}
\clearpage
\setcounter{page}{1}
\maketitlesupplementary
\setcounter{section}{0}
\renewcommand{\thefigure}{\arabic{figure}} 
\setcounter{figure}{0} 
\section{Data Collection}
\label{sec:Collection}
\subsection{Collection Devices}
\begin{figure}[htp]
\centering
\includegraphics[width=0.9\columnwidth]{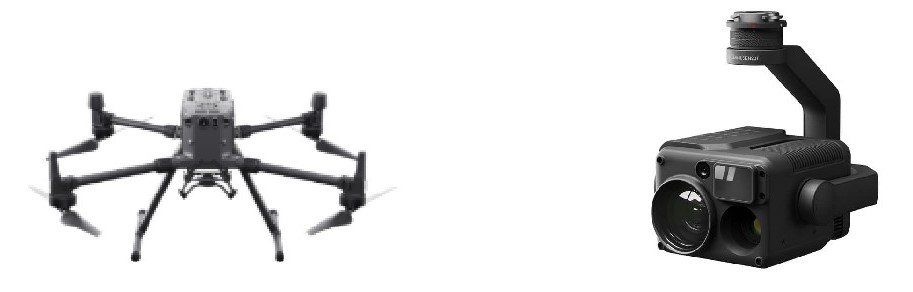} 
\caption{UVA platform and Cameras. DJI Matrice 300 RTK(Left) and DJI H20T(Right).}
\label{fig11}
\end{figure}

We use DJI products to collect the required data. Specifically, we employ the DJI Matrice 300 RTK as the drone platform, equipped with the DJI H20T camera to capture infrared data, as shown in Figure ~\ref{fig11}. The DJI Matrice 300 RTK is a professional commercial drone with a maximum flight time of 55 minutes and a horizontal positioning error of only 1 cm and vertical positioning error of 1.5 cm in areas with good signal quality. The H20T is a multi-sensor camera, and its long-wave infrared sensor has a temperature measurement range of -40°C to 150°C, which is sufficient to cover most scenarios in typical environments. The camera parameters of the H20T are shown in Table~\ref{tab:3}.
\begin{table}[h!]
\centering
\begin{tabular}{c|c}
\hline
\textbf{Parameter} & \textbf{H20T} \\ \hline
Sensor & Uncooled VOx Microbolometer \\ 
Image Resolution & 640 $\times$ 512 \\ 
Pixel Pitch & 12 $\mu m$  \\ 
Spectral Band & 8 $\mu m$  to 14 $\mu m$  \\ \
Focal Length & 13.5 mm \\ 
Temperature Range & -40°C to 150°C \\ \hline
\end{tabular}
\caption{H20T Camera Parameters}
\label{tab:3}
\end{table}

\subsection{Data Acquisition}
\begin{figure*}[!h]
\centering
\includegraphics[width=0.9\linewidth]{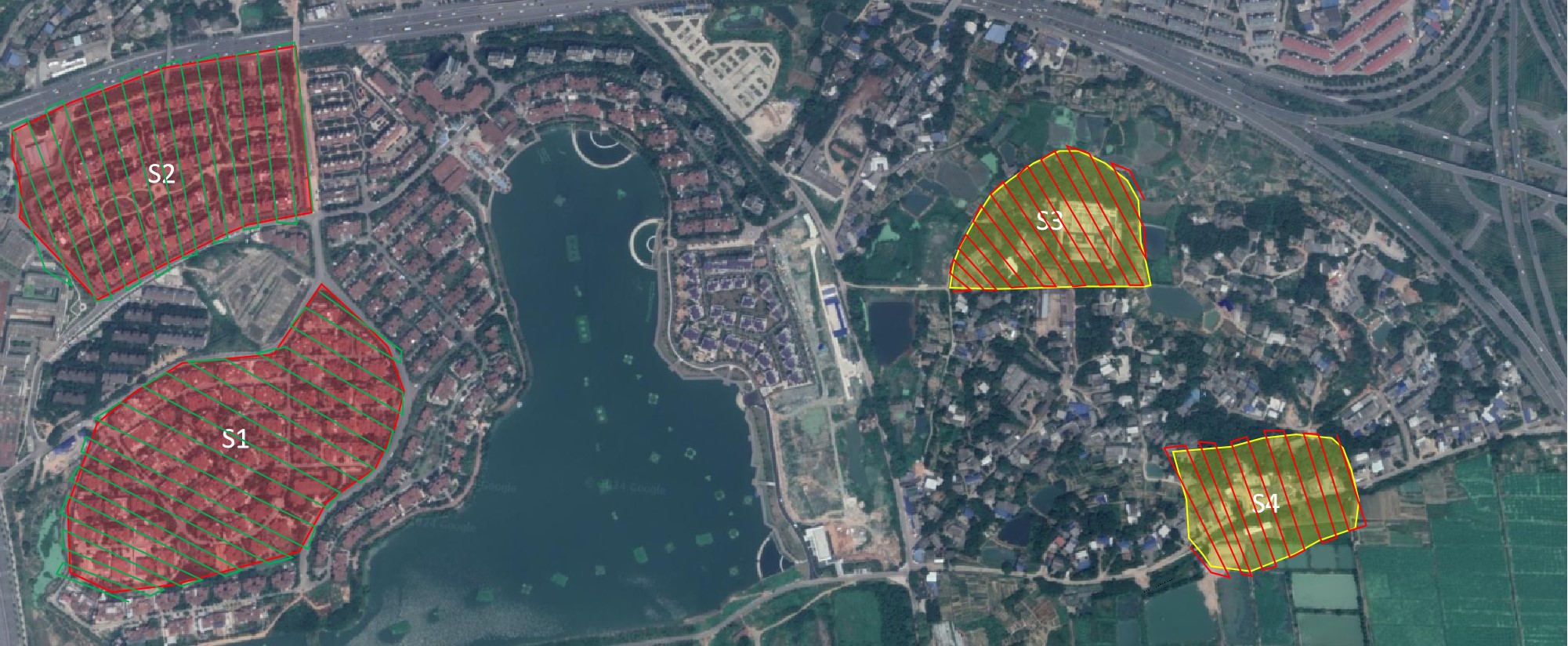} 
\caption{Area and route planning. }
\label{fig12}
\end{figure*}
The NTR dataset covers four regions, as shown in the Figure ~\ref{fig12}. To ensure the drone captures the entire area, we planned the flight path as illustrated, keeping the drone at a constant altitude of 250 meters above the ground and recording the time of capture during the flight. Since we aimed to collect nighttime data, the data collection started at sunset and continued until midnight. By utilizing the repeatability feature of the M300 RTK drone, we were able to fly the same flight path multiple times and capture thermal radiation data from the same location at different times, with intervals of approximately 2 to 3 hours. Additionally, factors such as GPS signal deviation and wind speed may cause slight positional offsets during the flight.

\subsection{Data Processing}
The data processing consists of two main parts: the generation of the initial point clouds  and the processing of thermal infrared images.

First, for the generation of the initial point cloud, we use a 3D textured model with absolute coordinates, created from high-resolution RGB images using photogrammetry software. This model is generated by sampling points from the mesh and performing random down-sampling to create a sparser initial point cloud, with approximately 300,000 points per scene. The reason for generating a sparse initial point cloud is that dense point clouds could lead to memory overflow issues during training.

Second, for the processing of thermal infrared images, the thermal infrared images captured by the H20T are in R-JPG format, containing basic information such as radiation intensity, capture time, and GPS location. However, the internal processing of the H20T applies various image enhancements, such as contrast stretching, to make the images more suitable for human visual perception. To ensure that the constructed 3D thermal radiation field accurately reflects temperature, NTR also provides the raw images recording radiation temperature. Using DJI's Thermal Imaging SDK (TSDK), we convert the R-JPG thermal infrared images into these raw images.

Furthermore, for accurate 3D reconstruction of the thermal infrared images, it is crucial to precisely estimate their absolute orientation. We adopt a rendering-based and matching-based pose estimation method , where the thermal image is registered to a pre-constructed RGB reference 3D model. The overall method includes three stages: (1) Rendering reference images and depth maps from the pre-constructed RGB 3D model using drone sensor priors. (2) Performing cross-modal feature matching between the rendered RGB reference image and the thermal image, establishing a 2D-3D correspondence. (3) Estimating the pose of the thermal image using the Perspective-n-Point (PnP) algorithm . Using the new pose, we render a synthetic RGB image from the high-resolution RGB model with absolute coordinates, which closely matches the thermal infrared image. These images are then used for feature generation with Feature-GS.
\section{Physical Quantity Analysis}
\label{sec:Analysis}
\begin{figure}[htp]
\centering
\includegraphics[width=0.9\columnwidth]{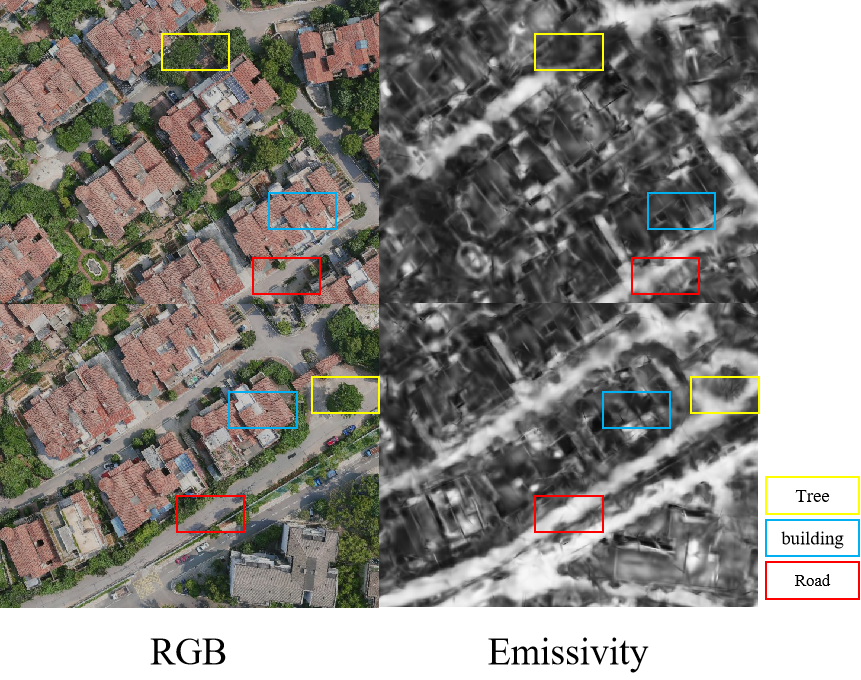} 
\caption{Taking emissivity as an example, we looked up the table to obtain the relative relationships among the buildings, roads and trees in the figure. They are roads > trees > buildings respectively. Therefore, our normalized emissivity map is relatively accurate and has obvious advantages over the artificial settings in traditional methods.}
\label{fig8}
\end{figure}
\begin{figure*}[htp]
\centering
\includegraphics[width=1.0\linewidth]{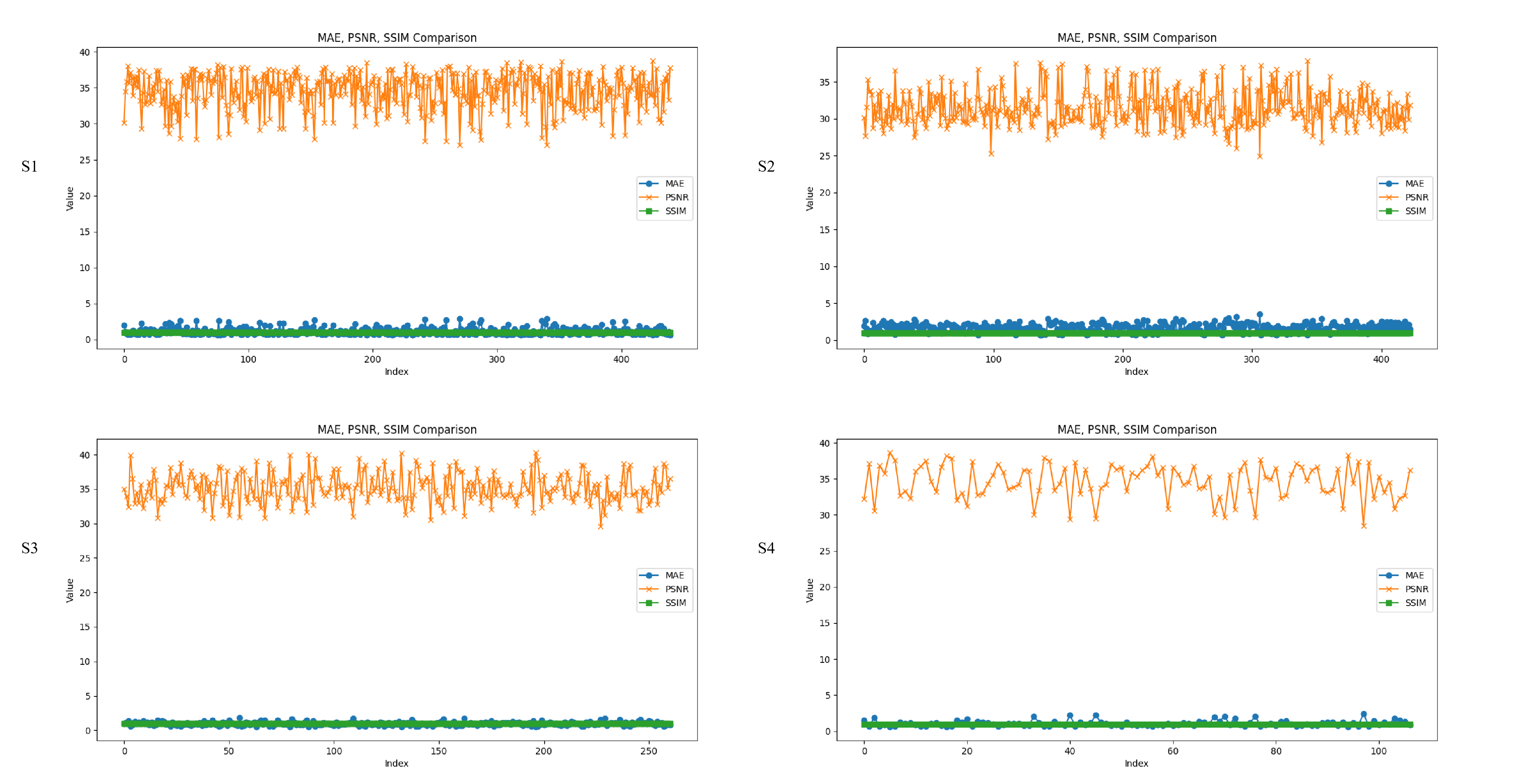} 
\caption{\textbf{Quantitative presentation of various scenes.} We presented the PSNR, SSIM and MAE of temperature between the predicted values and the ground truth values for each view in each scenario. Here, the abscissa represents the index of each view. It can be observed that our metrics for each view are relatively good.}
\label{fig9}
\end{figure*}
Our intermediate process will generate emissivity, convective heat transfer coefficient and heat capacity. Ours is different from the artificial settings. We obtain them through the network and then get the scene radiative temperature through numerical analysis, as shown in Fig.~\ref{fig8}. Taking emissivity as an example, we will demonstrate its accuracy. More results are in the NTR-Gaussian.MP4.
\section{More Experiment Results}
\label{sec:Results}
It is not sufficient to only present the average metrics. We take PSNR , SSIM and temperature MAE (Mean Absolute Error) as the evaluation metrics, and calculate these metrics under all the test views in areas S1, S2, S3 and S4 to illustrate the stability of our method. As shown in Fig.~\ref{fig9}, we presented the metrics of all views in all regions in the chart.

\begin{figure*}[!t]
\centering
\includegraphics[width=1.\textwidth]{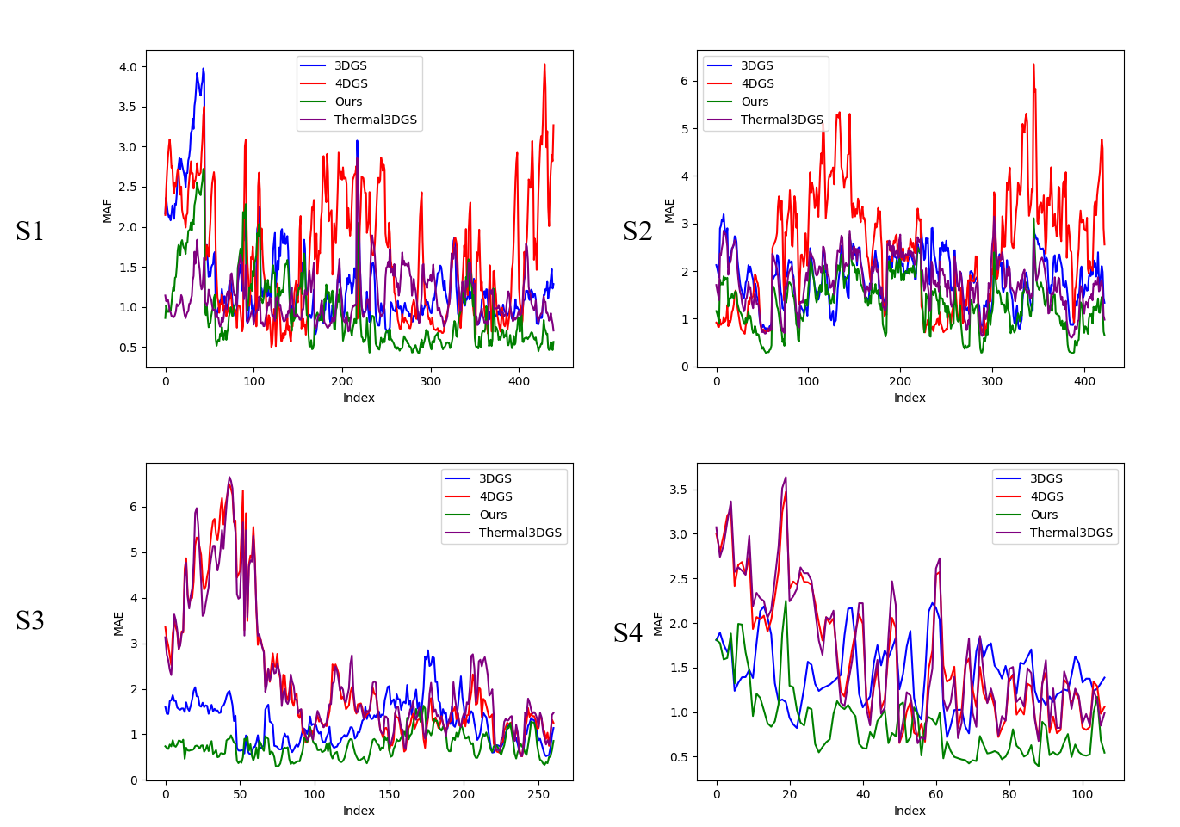} 
\caption{\textbf{Quantitative comparison of various scenes.} We presented the MAE of temperature between the predicted values and the ground truth values for each view in each scenario. Here, the abscissa represents the index of each view. It can be observed that our method has the lowest MAE among all the methods and outperforms the other methods.}
\label{fig10}
\end{figure*}

Taking the Mean Absolute Error (MAE) of temperature as a representative, we calculated the metrics of all methods in the test views of the four regions respectively to prove the superiority of our method. As shown in Fig.~\ref{fig10}, the MAE of our method is at the lowest level.

In addition, the dynamic temperature prediction under the new views can be seen in NTR-Gaussian.MP4.

\end{document}